\documentclass[conference]{IEEEtran}
\IEEEoverridecommandlockouts

\usepackage{cite}
\usepackage{amsmath,amssymb,amsfonts}
\usepackage{algorithmic}
\usepackage{graphicx}
\usepackage{textcomp}
\usepackage{xcolor}
\def\BibTeX{{\rm B\kern-.05em{\sc i\kern-.025em b}\kern-.08em
    T\kern-.1667em\lower.7ex\hbox{E}\kern-.125emX}}
\begin{document}

\title{Explainability and Continual Learning meet Federated Learning at the Network Edge}

\author{\IEEEauthorblockN{
Thomas Tsouparopoulos and Iordanis Koutsopoulos
}
\IEEEauthorblockA{Department of Informatics,
Athens University of Economics and Business\\
Athens, Greece \\ \textit{(Invited paper)}
}
}
\maketitle

\begin{abstract}
As edge devices become more capable and pervasive in wireless networks, there is growing interest in leveraging their collective compute power for distributed learning. However, optimizing learning at the network edge entails unique challenges, particularly when moving beyond conventional settings and objectives. While Federated Learning (FL) has emerged as a key paradigm for distributed model training, critical challenges persist. First, existing approaches often overlook the trade-off between predictive accuracy and interpretability. Second, they struggle to integrate inherently explainable models such as decision trees because their non-differentiable structure makes them not amenable to backpropagation-based training algorithms. Lastly, they lack meaningful mechanisms for continual Machine Learning (ML) model adaptation through Continual Learning (CL) in resource-limited environments. In this paper, we pave the way for a set of novel optimization problems that emerge in distributed learning at the network edge with wirelessly interconnected edge devices, and we identify key challenges and future directions. Specifically, we discuss how Multi-objective optimization (MOO) can be used to address the trade-off between predictive accuracy and explainability when using complex predictive models. Next, we discuss the implications of integrating inherently explainable tree-based models into distributed learning settings. Finally, we investigate how CL strategies can be effectively combined with FL to support adaptive, lifelong learning when limited-size buffers are used to store past data for retraining. Our approach offers a cohesive set of tools for designing privacy-preserving, adaptive, and trustworthy ML solutions tailored to the demands of edge computing and intelligent services. 
\end{abstract}



\section{Introduction}
The proliferation of Internet-of-Things (IoT) devices and the advent of edge computing have shifted the landscape of data generation and processing from centralized data centers to the network edge, where devices rely on wireless communication. This evolution calls for distributed learning strategies that can process data locally, preserve privacy, and overcome resource constraints \cite{Xia}. As a result, distributed learning paradigms, and most notably Federated Learning (FL), have emerged as promising solutions in these settings. Despite its potential, distributed learning introduces a variety of optimization challenges that must be addressed to harness its full benefits in dynamic, resource-limited settings.

\par Federated learning offers a compelling solution to distributed learning by enabling multiple wirelessly connected nodes to collaboratively train a shared model without sharing their data \cite{tsoup}. In the typical FL setup \cite{McMahan}, a server coordinates a distributed learning process where edge devices locally train a shared model using their private datasets. Each node performs model updates based on its local data and periodically transmits these updates to the server, where they are aggregated to refine the global model. This iterative process continues across multiple communication rounds until the model converges. Most FL approaches focus primarily on complex predictive models, and they often overlook the need for model interpretability and adaptability in the presence of streaming data and volatile network conditions.

\par  In addition to addressing privacy and resource limitations, \textit{modern applications and services require models that are both accurate and interpretable}. To meet these dual requirements, our work adopts two complementary approaches. First, when complex predictive models are used, we integrate Multi-objective Optimization (MOO) techniques to balance the objectives of predictive performance and explainability, using interpretable surrogate models to approximate their predictions \cite{Charal}. The challenge here lies in balancing the trade-off between these conflicting objectives while considering resource limitations and data privacy inherent to edge devices. Second, in scenarios where inherently interpretable tree-based models are needed, we discuss the difficulties of integrating them into a distributed learning setting. 

\par Furthermore, the adaptability of ML models to new tasks and data is a key pillar of ML, which is why Continual Learning (CL) is a critical component of distributed learning at the edge. Through CL, the system learns from new data, ensuring that the models can adapt without the need for computationally expensive retraining processes with all past data. We focus on replay buffer mechanisms within online CL scenarios and examine their limitations. When extending these techniques to FL, new challenges emerge as data remain partitioned across wirelessly connected devices and communication cost is influenced by network factors such as bandwidth limitations, latency, and congestion. 

\par In this paper, we investigate the 
optimization challenges underlying three critical and interrelated areas: Explainable Artificial Intelligence (XAI) with surrogate models, distributed decision-tree learning, and online Continual Learning with replay buffers. First, we develop rigorous formulations for each topic within a centralized framework, highlighting their respective performance trade-offs. We then extend these formulations to a distributed learning setting where clients are assumed to be wirelessly connected edge devices with limited resources, with a particular emphasis on FL, where data are partitioned across multiple clients and must remain private. 
\par Our unified view of distributed learning at the edge provides a cohesive and forward-looking strategy for building robust, adaptive, and trustworthy ML solutions tailored to the challenges of edge computing and intelligent edge services. While this work focuses on foundational optimization challenges that pervade wireless networks, it does not explicitly incorporate wireless-specific factors such as time-varying channel quality and node volatility. However, the proposed modelling framework is designed to be extensible, allowing for the integration of these aspects in practical implementations. The key contributions of our work are as follows:
\begin{itemize}

\item We introduce a MOO problem that jointly balances the predictive loss of a black-box model and the fidelity loss of an interpretable surrogate one.
\item We identify 
limitations in training inherently explainable decision trees across multiple clients in an FL setting. 
\item We investigate centralized and decentralized architectures in online CL with replay buffers, highlighting trade-offs in training coordination, efficient data transfer and communication overhead.
\item We develop rigorous formulations for the three interrelated areas of XAI with surrogate models, distributed decision-tree learning, and online CL with replay buffers, and extend them to the challenging FL setting.
\end{itemize}

\section{Related Work}
\par \textbf{Explainable AI with surrogate models.} Surrogate models have been used in the literature as a means to approximate the predictions of complex, black-box predictive models, and they are categorized into global and local surrogates. Global surrogate models \cite{Charal} aim at approximating the predictions of the black-box model on the entire dataset, while local surrogate models \cite{Ribeiro} are trained to provide explanations for individual instances of the dataset. Surrogate models are typically trained by minimizing the \textit{Point Fidelity} loss function \cite{Plumb}, which measures the disagreement between the output of the black-box model and the output of the surrogate one.

\par \textbf{Multi-objective learning.} When surrogate models are used to approximate, and therefore explain the decision making process of complex predictive models, the parameters of the black-box model benefit from being optimized concurrently for the predictive loss and \textit{Point Fidelity} \cite{Charal}, resulting in a MOO problem. While many prominent works \cite{Fei, Sener, Désidéri, Hu, Yang} have provided frameworks to find Pareto optimal solutions for MOO problems, they often rely on simplified assumptions about the environment and fail to fully capture the uncertainty and heterogeneity of real-world deployments at the network edge. Gradient-based MOO algorithms, such as the Multiple Gradient Descent Algorithm (MGDA) \cite{Désidéri} and its extensions \cite{Sener} are more suitable for large-scale problems, but their application in FL settings with strong theoretical guarantees remains underexplored, particularly in addressing data heterogeneity and conflicting objectives like model accuracy and explainability. Recent works, such as those in \cite{Hu, Yang}, have explored FL with multiple objectives, but have not integrated explainability or explicitly accounted for data heterogeneity. 
\par \textbf{Tree-based learning.} Existing approaches to federated decision-tree learning are relatively limited. Li et al. \cite{QLi} underscore that the non-parametric nature of decision trees precludes the direct use of gradient-based optimization, complicating their integration into distributed learning settings. Ludwig et al. \cite{Ludwig} propose a server-coordinated \textit{ID3} algorithm where clients compute local feature frequencies and send them to the server to make global split decisions. However, this necessitates frequent client-server communication, introducing scalability bottlenecks for deep trees or large client cohorts. Li et al. \cite{Patton} adapt AdaBoost for horizontal FL with weak decision tree learners, but the exchange of all weak learners (local decision trees) between all clients incurs privacy leakage risks and communication overhead. Garrido et al. \cite{Argente-Garrido} develop an aggregation process that merges decision paths from locally trained trees into a unified global model, but the aggregation mechanism struggles to ensure structural uniformity across heterogeneous datasets. These works highlight the fundamental trade-offs between privacy, communication efficiency, and model performance in distributed decision tree training. 


\par \textbf{Continual Learning} Continual learning approaches can be broadly categorized into three main paradigms: (\textit{i}) \textit{regularization-based methods}, which constrain updates to neural network (NN) parameters that are crucial for previously learned tasks \cite{Aljundi, Slin}; (\textit{ii}) \textit{parameter-isolation based methods} that learn which weights are important for old tasks to freeze during training with new tasks \cite{Yoon20}, or further expand the NN with new parameters when needed \cite{Yoon}; and (\textit{iii}) \textit{replay-based methods}, which either store and replay important data of old tasks when learning the new task \cite{Chrysakis, Nikol24} or store gradient information \cite{Lopez-Paz}.

\par \textbf{Continual Learning with replay buffers.} Replay-based methods \cite{Vitter, Chrysakis, Nikol24, Lopez-Paz} maintain a buffer of past training samples and interleave them with new data during training to maintain prior knowledge. For classification problems, the Reservoir Sampling ($\mathrm{RS}$) algorithm \cite{Vitter} stochastically stores a representative subset of the data stream
. Class-Balancing Reservoir Sampling ($\mathrm{CBRS}$) \cite{Chrysakis} extends ($\mathrm{RS}$) by enforcing class balance in the buffer, prioritizing under-represented classes. Kullback–Leibler Reservoir Sampling ($\mathrm{KLRS}$) updates the buffer content by minimizing the $\mathrm{KL}$ loss between the buffer distribution and a
target distribution dependent on the evolving empirical distribution of classes in the training data stream \cite{Nikol24}. Gradient Episodic Memory \cite{Lopez-Paz} combines replay with gradient constraints but requires task boundaries, limiting its applicability to online settings.

\section{Explainable AI with surrogate models \\in a Federated Learning setting}
\label{section:xai}
In safety-critical domains like healthcare, AI models must not only achieve high predictive accuracy but also offer transparency to ensure trust in their decision-making process. A common and efficient approach to achieve explainability is to use human-interpretable surrogate models to approximate the predictions of complex, high-accuracy models. In a mobile healthcare diagnostic system, FL enables multiple wirelessly connected devices to collaboratively train a global model for predictions and a surrogate one for explainability, without sharing sensitive data. However, training the two models separately results in a suboptimal trade-off between predictive performance and explainability. To this end, we propose a general MOO-based framework to address this trade-off and can be extended to accommodate network-related factors as additional loss terms, such as time-varying channel conditions and latency, depending on the application.

\subsection{A centralized approach with Multi-objective Optimization for Explainable AI}

In a centralized setting with all data in one location, consider a black-box model \( f_{\theta}:\mathbb{R}^d \to \mathbb{R} \), parameterized by \( \theta \), which is trained to perform a supervised learning task, such as classification or regression, where $d$ is the dimension of input data. Alongside this model, an interpretable surrogate model \( g_{\phi}: \mathbb{R}^d \to \mathbb{R} \) parameterized by \( \phi \), such as a linear regression one, is optimized to approximate the black-box model's predictions.

\par Given a dataset \( \mathcal{D} = \{(x_i, y_i)\}_{i=1}^{N} \), the problem is captured by two objectives. The first objective is to minimize a typical prediction loss function, 
\begin{equation}
L_{\text{pred}}(\theta) = \frac{1}{N} \sum_{i=1}^{N} \ell\big(f_{\theta}(x_i), y_i\big),
\end{equation}
where \( \ell(\cdot, \cdot) \) represents a task-specific loss function such as cross-entropy or mean squared error. The second objective ensures that the surrogate model accurately approximates the black-box model by minimizing the difference between their predictions, expressed through the \textit{Point Fidelity} ($\mathrm{PF}$) loss:
\begin{equation}
L_{\text{PF}}(\theta, \phi) = \frac{1}{N} \sum_{i=1}^{N} \big( g_{\phi}(x_i) - f_{\theta}(x_i) \big)^2.
\end{equation}

\par Optimizing the parameters of the black-box model simultaneously for these two objectives is crucial for maintaining both high predictive performance \textit{and} model interpretability \cite{Charal}. However, there is an inherent trade-off between these two objectives, as improving explainability (as captured by $\mathrm{PF}$ loss), by forcing the black-box model to be more aligned with the surrogate model, may come at the expense of reducing the predictive performance of the black-box model. 
\par This trade-off can be addressed by formulating the problem as a bi-level optimization problem \cite{Charal25} wherein the upper-level is the MOO problem that optimizes the parameters of the black-box model for the two objectives and the lower-level problem optimizes the parameters of surrogate one to best approximate the black-box model with fixed parameters:
\begin{align}
\label{Problem_1}
\min_{\theta} \quad & \Big[ L_{\text{pred}}(\theta), L_{\text{PF}}(\theta, \phi^*) \Big], \\
\text{s.t.} \quad & \phi^* \in \arg\min_{\phi} L_{\text{PF}}(\theta, \phi). \nonumber
\end{align}
Intuitively, this means that the joint objective manages to ``confine'' the black-box model's parameters so that they can be better approximated by the surrogate one, without compromising the black-box model's predictive performance.

\subsection{Federated Multi-objective Optimization}

In high-stakes distributed learning settings, such as healthcare, data are distributed across geographically dispersed clients. For example, in an FL setting, several hospitals wish to collaboratively train a shared model to predict the occurrence of a side effect while preserving patients' data privacy, and at the same time use a surrogate model to explain it by approximating its predictions. 
\par However, the deployment of ML models in such contexts is impeded by the need for both high predictive accuracy and explainability in decision making. In the centralized case, a single model is optimized on the whole dataset, and MOO can directly the conflicts between objectives using, for example, gradient-based approaches \cite{Désidéri}. In contrast, the goal in FL is to obtain a single global solution that balances the trade-off between predictive accuracy and explainability given that different clients have separate datasets which are usually statistically heterogeneous.

In this FL setting with $M$ clients, each client \( k \) has access to its own dataset \( \mathcal{D}_k = \{(x_i^{(k)}, y_i^{(k)})\}_{i=1}^{N_k} \) and seeks to collaboratively train a shared pair of models: a black-box model \( f_{\theta} \) and a surrogate model \( g_{\phi} \). The problem formulation is thus extended in an FL setting as follows:

\begin{align}
\min_{\theta} \quad &  \Bigl[\sum_{k=1}^{M} L_{\text{pred}}^{(k)}(\theta), \sum_{k=1}^{M} L_{\text{PF}}^{(k)}(\theta, \phi^*) \Bigr], \\
\text{s.t.} \quad & \phi^* \in \arg\min_{\phi}  \sum_{k=1}^{M} L_{\text{PF}}^{(k)}(\theta, \phi), \nonumber
\end{align}
where each client \( k \) optimizes its own predictive loss 
\begin{equation}
L_{\text{pred}}^{(k)}(\theta_k) = \frac{1}{N_k} \sum_{i=1}^{N_k} \ell\big(f_{\theta_k}(x_i^{(k)}), y_i^{(k)}\big),
\end{equation}
and its own $\mathrm{PF}$ loss 
\begin{equation}
L_{\text{PF}}^{(k)}(\theta_k, \phi_k) = \frac{1}{N_k} \sum_{i=1}^{N_k} \big( g_{\phi_k}(x_i^{(k)}) - f_{\theta_k}(x_i^{(k)}) \big)^2.
\end{equation}

\subsection{Future directions}
\subsubsection{Addressing the accuracy-explainability tradeoff in an FL setting} 
Despite the theoretical guarantees that MOO methods provide, novel MOO algorithms must be developed to better balance conflicting objectives such as predictive accuracy, interpretability, and resource utilization in FL settings. In resource-constrained edge environments, computing the exact Pareto front for MOO problems is computationally intractable, making it impractical to address all objectives simultaneously. This challenge motivates the development of real-time approximation methods endowed with theoretical guarantees on convergence. In such settings, surrogate models serve as a key tool by approximating complex models and thereby reducing computational complexity. In particular, efficient gradient-based and surrogate-model-assisted techniques that approximate the Pareto front in the presence of heterogeneous data are a promising research direction. 

\subsubsection{Federated Learning of personalized explainable models}
In a realistic FL scenario, clients typically have heterogeneous datasets, i.e., local client's data are drawn from different distributions. For instance, in healthcare, each hospital maintains its own dataset, with unique demographic and clinical features. This environment presents significant challenges for training accurate and explainable models with MOO due to data heterogeneity, privacy constraints, and decentralized optimization, making it difficult to find a single Pareto optimal solution that performs well on all clients' data. This heterogeneity necessitates personalized solutions tailored to each client’s local data, as their Pareto fronts for conflicting objectives (e.g., predictive accuracy vs. \textit{Point Fidelity}) differ. To attain this goal, each client could seek to train its own pair of personalized models (a local black-box model and a local surrogate model) by solving the MOO problem locally.

\section{Decision-tree Learning}
\label{sec:dts_fl}

\par
\begin{figure}[t!]
\centering
\includegraphics[scale=0.138]{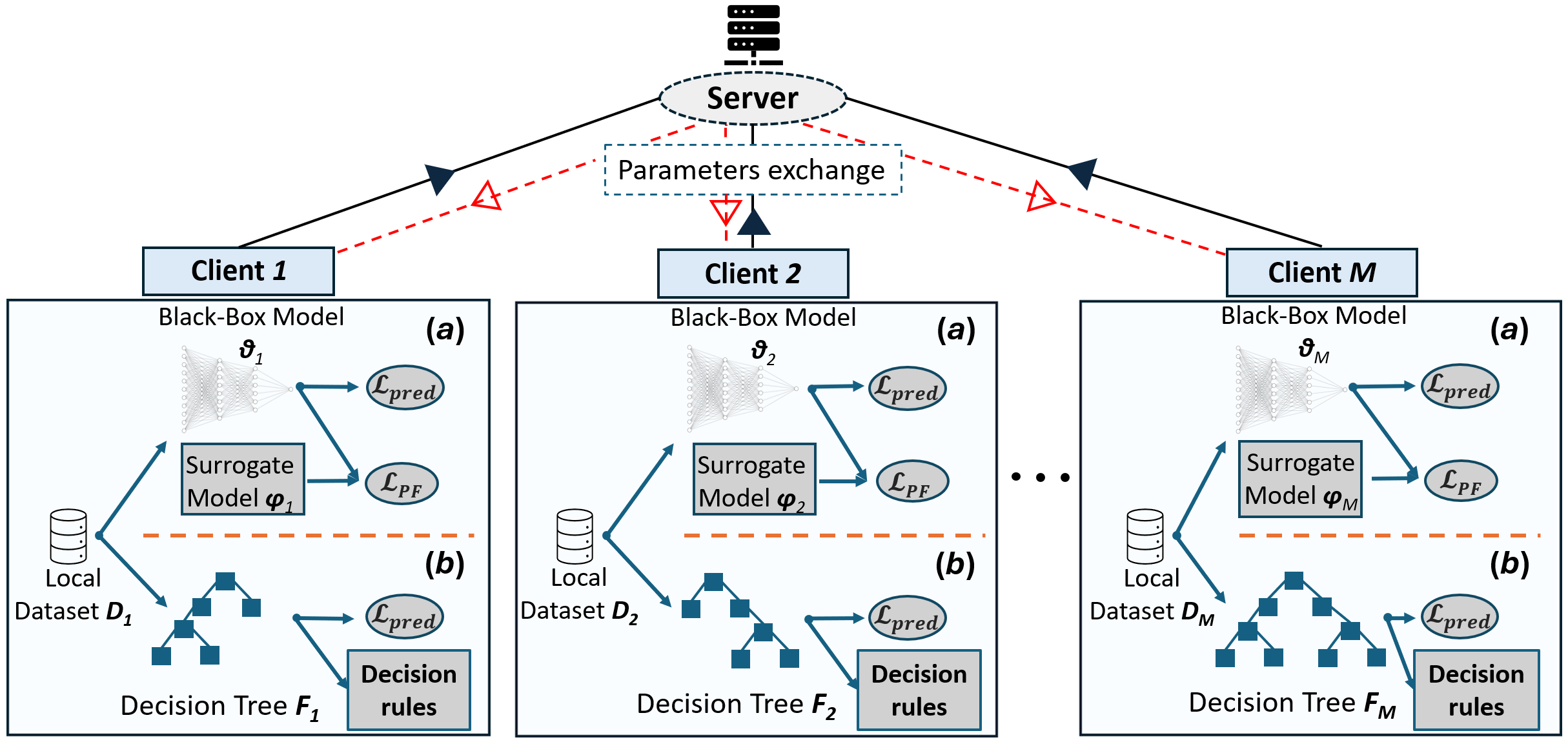} 
\caption{A Federated Learning (FL) setting, where explainability is attained either with a surrogate model explaining a black-box predictive model e.g. a linear regression model (upper part, (\textit{a}) of the three small figures), or with a self-explainable decision tree (bottom part, (\textit{b}) of the three small figures).} \label{fig:FL_XAI_MOO}
\end{figure}

 Decision trees (DTs) are non-parametric hierarchical models that play a crucial role for XAI in a Federated Learning (FL) setting. Unlike differentiable surrogate models that approximate the predictions of complex models to enhance transparency, DTs offer inherent interpretability through human-readable decision rules, eliminating the need for additional explainability mechanisms. This inherent transparency, combined with their low computational cost and ability to handle both numerical and categorical data, makes DTs particularly well-suited for FL systems operating over wireless networks with resource-limited devices, where communication constraints and resource limitations are key considerations \cite{Rokach}. In Fig.~\ref{fig:FL_XAI_MOO} we illustrate how explainability meets FL at the network edge with two approaches: (\textit{i}) employing surrogate models to explain complex predictive models and (\textit{ii}) leveraging self-explainable decision trees, which offer a lightweight, interpretable alternative ideal for distributed, bandwidth-limited environments.

\subsection{Decision Tree training fundamentals}  
DTs make predictions  by recursively partitioning the input feature space into disjoint regions based on binary conditions defined by feature thresholds. Each intermediate node in the tree represents a decision rule in the form of binary conditions 
(e.g., ``\textit{Feature $x_j \leq$ threshold $\theta_j$}"), while leaves represent the final predictions. The training goal is to define intermediate nodes and find the optimal feature and corresponding threshold at each intermediate node by minimizing an \textit{impurity} measure, for instance the $\mathrm{Gini}$ impurity for classification or the mean squared error for regression tasks \cite{Rokach}. When training is complete, each path from the root to a leaf represents a human-readable prediction rule which is formed by combining logical AND with the binary conditions of intermediate nodes encountered along the path. This structure is a key strength of decision trees, since unlike complex models, their predictions can be easily understood through these human-readable rules, making them inherently explainable.

\par Formally, let $\mathcal{F} : \mathbb{R}^d \to \mathcal{Y}$ be a decision tree model, where $d$ is the dimension of the input feature space and $\mathcal{Y}$ is the output space, e.g., $\mathbb{R}$ for regression tasks and $\{1, \dots, C\}$ for classification tasks. Let also $\{(\mathbf{x}^{(i)}, y^{(i)})\}_{i=1}^{n}$ be a training set, where $\mathbf{x}^{(i)} = ({x}_1^{(i)}, \dots, {x}_d^{(i)}) \in  \mathbb{R}^d$ and  $y^{(i)} \in \mathcal{Y}$. The DT implements predictions by making partitions of the input domain $\mathbb{R}^d$ into $K$ disjoint regions $\{R_j\}_{j=1}^K$ satisfying $\bigcup_{j=1}^K R_j = \mathbb{R}^d$ and $R_i \cap R_j = \emptyset \ \forall i \neq j$. The prediction function for a given input $\mathbf{x} \in \mathbb{R}^d$ is then expressed as:
\begin{align}
\mathcal{F}(\mathbf{x}) = \sum_{j=1}^K w_j \cdot \mathbf{1}[\mathbf{x} \in R_j],
\end{align}
where $\mathbf{1}[\cdot]$ denotes the indicator function and $w_j$ represents the prediction for region $R_j$, such that every point in $\mathbb{R}^d$ belongs to exactly one region. In regression settings, $w_j$ is typically chosen as the average of the target values of the training data in $R_j$, whereas in classification, $w_j$ is the most frequent class label in $R_j$. 
\par The training objective is to minimize the empirical risk:
\begin{align}
L(\mathcal{F}) = \sum_{i=1}^{n} \sum_{\mathbf{x}^{(i)} \in R_j} \ell\bigl(y^{(i)}, w_j\bigr),
\end{align}
where $\ell(\cdot,\cdot)$ is a task-specific loss function. Unlike the cases of differentiable models discussed in section \ref{section:xai}, this is not a smooth loss function due to the discrete structure of the decision tree itself. Thus, the direct minimization of $\mathcal{L}(\mathcal{F})$ is computationally intractable due to the combinatorial nature of region selection. To this end, a greedy recursive partitioning strategy is necessitated.

At each node $S$ of the tree that contains a subset of the training data $D_S = \{(\mathbf{x}^{(i)}, y^{(i)})\}_{i=1}^{n_S}$, the greedy algorithm selects a feature $x_j \in \{x_1, \ldots, x_d\}$ and a threshold $\theta_j \in \mathbb{R}$ to maximize the impurity gain, based on a task-dependent cost function $C(\cdot)$ that is usually called impurity measure,
\begin{align}
G(x_j, \theta_j) = C(S) - \left( \frac{|\mathcal{L}|}{|D_S|} C(\mathcal{L}) + \frac{|\mathcal{R}|}{|D_S|} C(\mathcal{R}) \right),
\end{align}
where $\mathcal{L} = \{(\mathbf{x}, y) \in D_S : x_j \leq \theta_j\}$ and $\mathcal{R} = D_S \setminus \mathcal{L}$ denote the resulting partitions, and $|D_S|$, $|\mathcal{L}|$, $|\mathcal{R}|$ denote the number of examples in the corresponding sets. For a regression task, the impurity measure is typically the mean squared error:
\begin{align}
C(S) = \frac{1}{n_S} \sum_{i=1}^{n_S} \left(y^{(i)} - \bar{y}_S\right)^2,
\end{align}
where \(\bar{y}_S = \frac{1}{n_S} \sum_{k=1}^{n_S} y^{(k)} \) is the sample mean of the target values at node $S$. For classification, given class probabilities \( \{p_c\}_{c=1}^{C} \) where \( p_c \) represents the proportion of examples in \( S \) belonging to class \( c \), i.e. \(p_c = \frac{1}{n_S}\sum_{i=1}^{n_S}\mathbb{I}\{y^{(i)} = c\}, \,\,\, c=1,\dots,C.\), the impurity is often measured using entropy:
\begin{align}
C(S) = -\sum_{c=1}^{C} p_c \log p_c\,.
\end{align}

\par Once the optimal split is identified, the current node is partitioned accordingly, and the same procedure is recursively applied to each child node. This process continues until a stopping criterion is met, such as when the number of examples in a node falls below a pre-specified threshold, when a maximum tree depth is reached, or when the improvement from further splits becomes negligible. Leaf nodes assign $w_j$ as the mean target value (in regression) or majority class (in classification) of their respective subsets.

\par An important observation is that, although the threshold $\theta$ is a continuous parameter, it suffices to consider only a finite set of candidate thresholds. For a given feature $x_j$, the values $\{x^{(i)}_j\}_{i=1}^n$ can be sorted in non-decreasing order, and any threshold chosen between two successive distinct values results in the same partition of the training data. Thus, it is common practice to consider thresholds of the form \( \theta = (x^{(i)}_j + x^{(i+1)}_j)/2, \) thereby reducing the search over $\theta$ to a finite set of possibilities, i.e., $d \times (n_S - 1)$ splits.

\subsection{Decision Tree training with Federated Learning}

DTs exemplify the need for efficient, privacy-preserving algorithms that retain model interpretability---a cornerstone of trustworthy learning at the edge. In an FL setting, data are distributed across \( M \) clients, each holding a private dataset \( D_m \) drawn from a distribution 
$P_m(\mathbf{x}, y)$. 
DTs are inherently self-explainable because their predictions follow step-by-step rules based on feature splits, making it easy for humans to understand how a decision was made. 

\subsubsection{Training a single decision tree in an FL setting}
Training a single decision tree in an FL setting is challenging due to its non-differentiable structure, which complicates the use of gradient-based training algorithms. Decision-tree model training relies on identifying split-threshold pairs based on local data. This sequential and discrete training procedure prevents a straightforward means of aggregating the local splits. In contrast, federated ensembles like random forests or gradient-boosted trees circumvent these issues by design. 

\subsubsection{Training Tree Ensembles in an FL setting}
\par When training tree ensembles in an FL setting, each client can independently train multiple decision trees on its local dataset, and then contribute these trees to a global ensemble model. A specific example of this approach is \textit{Federated Random Forests} \cite{Argente-Garrido}, where each client constructs several decision trees in parallel and then a subset of trees from each client is selected and aggregated into a global ensemble model $M_{\text{global}} = \bigcup_{k=1}^M \mathcal{M}^k$, where for each client \( k \), \(\mathcal{M}^k\) is the set of trees chosen with \( |\mathcal{M}^k| = m_k \) (i.e., \( m_k \) trees are selected for client \( k \)). These trees can be chosen by random sampling or by using selection criteria based on a performance metric, e.g., favoring trees with low loss on held-out validation data with the goal of enhancing the overall ensemble performance.

\par \textit{Gradient Boosting} is a ML technique that builds a strong predictive model by combining many simpler models, known as \textit{weak learners} (typically decision trees). These trees are trained one after the other in a sequential process so that each new tree is trained to correct the residual errors made by the combined predictions of all previous trees. Similarly, \textit{Federated Gradient Boosting} \cite{Patton} extends this concept to a FL setting, where the goal is for clients to collaboratively train an ensemble without sharing their data. In this framework, each client sequentially trains trees to address residual errors made by the previous clients. After the local training phase, the models are securely transmitted to a central server. The server aggregates these locally trained ``weak" models to form a more accurate global predictor.

This flexibility in training multiple individual trees across clients in ensemble methods allows for a more straightforward implementation in distributed environments, where each tree can be trained separately and in parallel on different clients before being aggregated into a global model. This property simplifies parallelization and model aggregation at the server. However, despite their benefits, ensemble methods tend to be significantly more computationally demanding and less interpretable than single-tree models.

\subsection{Future directions}


In an FL setting, training tree-based models is sensitive to data heterogeneity among clients, which impedes the construction of a single one-fits-all DT model. \textit{Constrained tree growth strategies} can be designed to regulate how trees evolve at each client and ensure structural homogeneity. One such approach is to align local splits with a shared \textit{global feature partitioning scheme} whereby all clients follow a common guideline for selecting feature-threshold pairs when splitting nodes. By adhering to a shared global feature partitioning scheme, local trees can evolve within a common structured framework, thus minimizing structural discrepancies and promoting a coherent global model. Furthermore, online splitting criteria can be employed to synchronize local decisions in real time, further enhancing the consistency of local decisions among clients.

\section{Continual Learning with replay buffers}

In many real-world applications, the deployed ML models must continuously adapt to newly arriving data, without forgetting previously acquired knowledge. This challenge is particularly evident at the network edge where data arrive in streams or are generated in a continuous fashion. Unlike traditional static learning paradigms, ML models are required to make real-time updates in the presence of limited memory and computational resources. One common strategy to mitigate forgetting is the use of replay buffers, which store important and representative past seen examples to reinforce prior knowledge while training with new examples. However, managing these buffers presents several challenges, including memory constraints, data imbalance, and computational trade-offs. 

\subsection{Online Continual Learning with a replay buffer}

The setting of online Continual Learning involves a learner receiving a (potentially infinite) stream of training data in discrete time steps. At each time step $t$, a learner obtains a batch of training data $\mathcal{D}_t = \{(x_{i,t}, y_{i,t})\}_{i=1}^{N_t}$, where each input $x_{i,t}$ belongs to an input space $\mathbb{R}^d$ and each label $y_{i,t}$ is one of $K$ possible classes (supervised learning). Each learner maintains a replay buffer that keeps a subset of data points $\mathcal{B}$; hence the buffer capacity is $B = |\mathcal{B}|$. Replay buffers aim to address the issue of catastrophic forgetting by storing and replaying past data samples during training. 

\par 
The purpose of training in online Continual Learning is to update the model parameters $\theta$ at each step $t$, by optimizing a combined loss function $L(\cdot)$ that depends on both the current training batch $\mathcal{D}_t$ and the replay batch $\mathcal{B}_t$. Let \(\theta\) denote the model parameters and \(f(x;\theta)\) the prediction function (e.g., a deep neural network with softmax output). For any fixed dataset \(\mathcal{D} = \{(x_i,y_i)\}_{i=1}^{N}\), the empirical loss is defined as \(L_{\mathcal{D}}(\theta) = \frac{1}{N}\sum_{i=1}^{N}\ell\bigl(f(x_i;\theta), y_i\bigr),\) where \(\ell(\cdot,\cdot)\) is a task-specific loss function (e.g., cross-entropy for classification). The overall loss at time \(t\) is a convex combination of the loss on the current batch and that on the replayed data:
\begin{align}
\label{CL_eq_1}
L_t = \beta L_{\mathcal{D}_t} + (1 - \beta) L_{\mathcal{B}_t},
\end{align}
with \(\beta \in [0,1]\) balancing the weight placed on new and replayed data. Note that $\beta$ may also be varying with time.

\par Data arrive in (potentially infinite) streams of data $\{\mathcal{D}_1, \dots, \mathcal{D}_t, \dots\}$. For simplicity let us assume that $|\mathcal{D}_t| = B$ for all $t$. At $t=1$, the \( B \) arriving data points are stored in the buffer. 
For each $t>1$, when a new batch of training data \( \mathcal{D}_t \) arrives, the algorithm must decide how to update the buffer. For each data point in the arriving new batch, there exist the following two options: (\textit{i}) insert the data point into the buffer and remove from the buffer another data point (possibly from the same class), and (\textit{ii}) do nothing and ignore this data point. Once the buffer update is completed, the model is retrained using the combined dataset \( \mathcal{D}_t \cup \mathcal{B}_t \), leading to a new model \( \theta_{t+1} \). The updated model is then deployed to handle inference requests generated at time step \( t+1 \), ensuring continuous adaptation to the evolving data stream while maintaining a buffer with representative data.

\par In online continual learning, non-stationary data distribution refers to the phenomenon where the empirical class distribution, call it $p(t)$, is not fixed but evolves over time. This means that the relative frequencies of the $K$ classes (in the case of classification tasks) change as new data arrive, 
leading to fluctuating proportions of class examples in the stream. This non-stationary nature complicates buffer management. If the buffer simply mirrors the natural imbalances present in the incoming stream, classes that are infrequently observed may become severely underrepresented, degrading the model's ability to retain information about them. Conversely, if the buffer is forced to maintain a balanced representation of classes, it might no longer accurately reflect the true, evolving distribution of the data, potentially leading to misaligned learning outcomes. 

There exist recent approaches that adjust the stored distributions using divergence measures like the Kullback-Leibler loss to tailor the buffer to better approximate a desired target distribution \cite{Nikol24}. 
This balance between fidelity to the evolving data stream and the need for fair (in terms of classes) and efficient learning underscores the inherent complexity of Continual Learning systems at the network edge and invites a forward-thinking exploration of adaptive, efficient strategies to manage non-stationary data.


\subsection{Federated Continual Learning}

\begin{figure}[t!]
\centering
\includegraphics[scale=0.165]{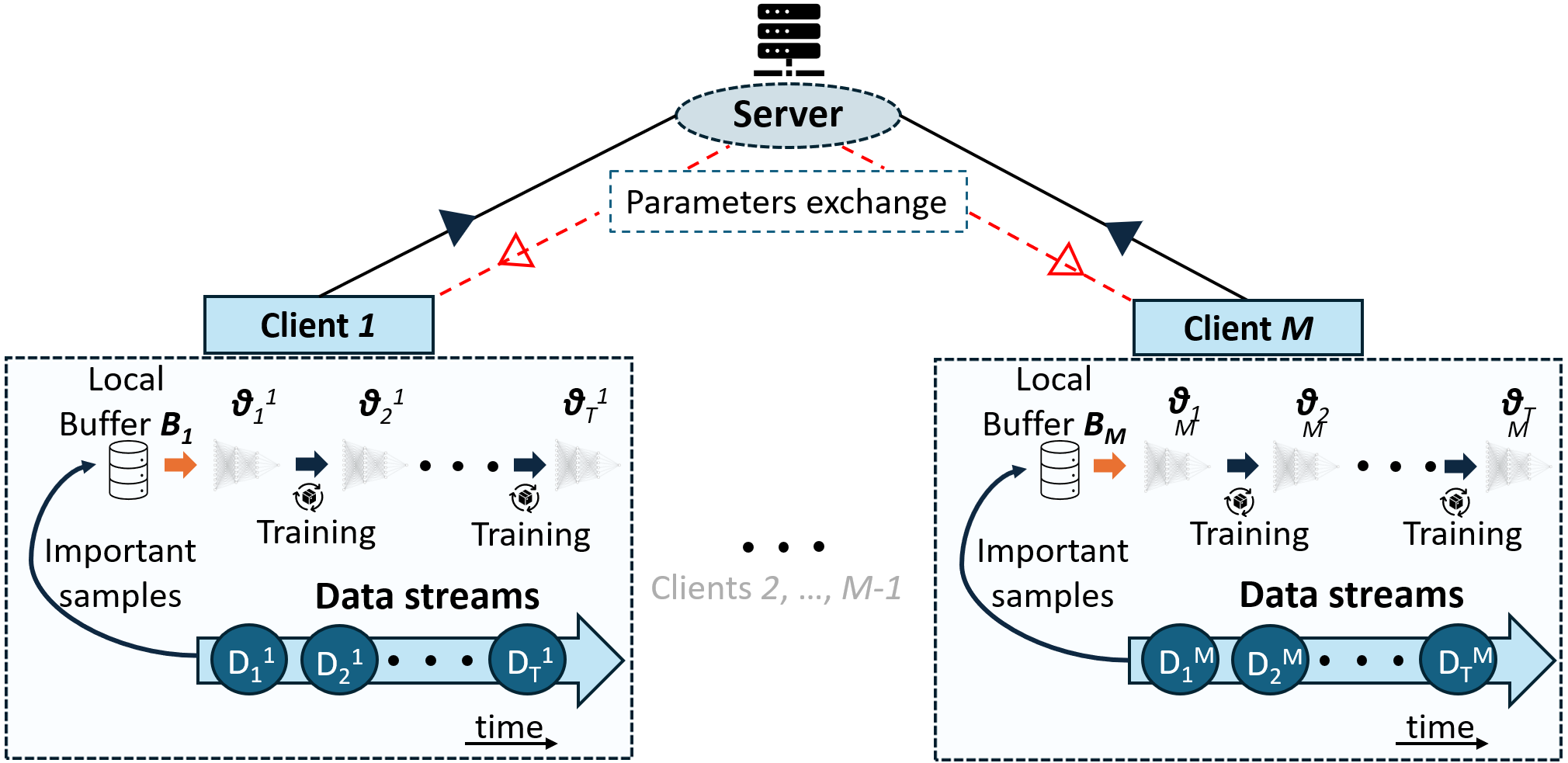} 
\caption{A Federated Continual Learning (FCL) framework wherein each client has its own buffer, and new data arrive in streams. The local models evolve by continuously adapting to new local data and feedback from the aggregator.} \label{fig:FCL}
\end{figure}

In certain cases, the requirement to have a continually updated model at each time aligns with the reality of having distributed data at different clients that cannot be moved to a central location. For example, consider a set of hospitals, each with its own dataset. A model must be trained on data from all hospitals and kept continuously updated. Privacy concerns prevent clients from sharing their local data, particularly when this involves sensitive information like patient records. Clients collect new data continuously, and models must be retrained  to account for new information such as variations in patient demographics or various emerging trends that cause distribution shifts in data. Federated Continual Learning (FCL) (shown in Fig.~\ref{fig:FCL}) addresses this challenge by enabling multiple clients to collaboratively train and update a single model $\theta_t$ at each time $t$, ensuring that it remains up-to-date and responsive to local inference requests at each client site \cite{Yoon21}. 

\par In a FCL setting, each client $i$ receives a non-stationary batch of training data $\mathcal{D}_t^{(i)} = \{(x_{j,t}^{(i)}, y_{j,t}^{(i)})\}_{j=1}^{N_t^{(i)}}$ and maintains its own local buffered subset of data $\mathcal{B}^{(i)}$ of capacity $B_i = |\mathcal{B}^{(i)}|$. Local model updates are performed by minimizing equation~\eqref{CL_eq_1}, and these updates are periodically aggregated at a server using methods such as FedAvg \cite{McMahan}, to form a global model:
\begin{align}
    \theta_{t+1} = \sum_{i=1}^{M} w_t^{(i)} \theta_t^{(i)},
\end{align}
where $w_t^{(i)}$ is a weighting factor for each client. 
\par One of the key challenges in FCL with local replay buffers is the non-stationary data processes across clients. Since each client has access to a different subset of the global data distribution, local buffers may not capture sufficient diversity since they over-represent client-specific classes, leading to poor performance of the global model on unseen data. Privacy constraints limit the possibility of sharing buffer contents between clients, making it difficult to establish a unified representation of past data that would resemble the global distribution of data. 

\subsection{Distributed Continual Learning}

When extending Federated CL to a distributed setting with multiple clients with the possibility of data exchange across clients, the model becomes ever richer. 
Distributed CL can be formulated as an optimization task that aims to balance continual model accuracy with computational and communication efficiency. The network between clients are represented as a graph $G = (\mathcal{V}, \mathcal{E})$ where $\mathcal{V}$ represents the set of $M$ clients and $\mathcal{E}$ denotes the communication links between them. Each client $i$ at time $t$ generates or collects a dataset $\mathcal{D}_{i,t}$, and depending on the architectural paradigm, it either maintains a local replay buffer with data $\mathcal{B}_{i,t}$, or a global buffer with stored data $\mathcal{B}_t$ exists at a server. 
\par The objective is as follows:
\begin{equation}
\min_{\{\mathcal{D}_{i,t}^{\text{proc}}, \mathcal{D}_{i,t}^{\text{trans}}, \mathcal{D}_{i,t}^{\text{rec}}\}} \sum_{t=1}^{T} \sum_{i=1}^{M} \lambda L_i(\mathcal{D}_{i,t}^{\text{proc}}) + (1-\lambda) \text{Cost}(\mathcal{D}_{i,t}^{\text{trans}}, \mathcal{D}_{i,t}^{\text{rec}}),
\end{equation}
where $\lambda \in [0,1]$ is a trade-off parameter between the loss incurred from the data processed at each client $i$, $\mathcal{D}_{i,t}^{\text{proc}}$, and the cost incurred from data transferred to other clients, $\mathcal{D}_{i,t}^{\text{trans}}$ and received by client $i$ from other clients, $\mathcal{D}_{i,t}^{\text{rec}}$. Function $L_i(\cdot)$ represents the training loss as defined in equation \eqref{CL_eq_1}, and $\text{Cost}(\cdot, \cdot)$ quantifies the cost of communication overhead for the network, e.g. in terms of energy or delay. The optimization problem is subject to constraints which ensure that data used for training are within the available local and received datasets, they flow in the network subject to link capacity constraints, and they adhere to memory constraints of the replay buffer and node computational capacity constraints. 

The placement of replay buffers also plays an important role. Two examples of architectural paradigms for deploying replay buffers are: (\textit{i}) a single shared buffer placed at a trusted node that is accessible by all clients, or (\textit{ii}) multiple local buffers, each deployed at each client. Each approach presents different trade-offs between collaboration efficiency, resource utilization, and personalization. A single shared buffer simplifies coordination but risks communication overhead and stale data, while multiple local buffers enable task-specific adaptation at the cost of complicated decentralized management. The primary challenge in both scenarios is the limited buffer capacity, which forces each local learner to either discard data points or transfer them to its neighbors, potentially leading to performance loss or communication overhead. 

\subsection{Future directions}

\subsubsection{Distributed CL with many replay buffers}

\par According to one approach in distributed CL, each client \(i\) can maintain a local replay buffer $\mathcal{B}_i$ tailored to its data distribution. 
Clients can periodically exchange samples from their buffers with neighbors or with a central aggregator. 
In this setting, each client \(i\) selects a subset of data $\mathcal{D}_{i,t}^{\text{proc}}$ to process for training, which consists of samples from its current stream of data, its local buffer, and potentially data received from other clients to train on. Additionally, the client determines the subset of data $\mathcal{D}_{i,t}^{\text{cache}}$ to retain in its buffer and the subset of data $\mathcal{D}_{i,t}^{\text{trans}}$ to transfer to other clients. 

\par This approach would require decentralized data routing. Clients would need to decide which data to retain, transfer or discard based on network topology and their resource availability. 
For instance, a client with limited computational capacity might offload excess data to neighboring devices, incurring a multi-hop communication cost. The time-coupled nature of the problem means that current data caching decisions influence future model performance and resource use. 

\subsubsection{Distributed CL with one replay buffer}

Alternatively, a central replay buffer at a trusted server node is shared among all clients and aggregates a subset of important prior data samples from the entire network of clients.
Centralized coordination requires synchronization between the server and the client nodes and introduces communication overhead, as nodes must transmit data to and from the shared buffer. This centralized coordination also assumes reliable connectivity to the buffer (e.g., a server), which may not always exist. Another key challenge in this setting lies in maintaining \textit{data freshness}. Overly frequent data replacements risk discarding critical data for past knowledge retention, while infrequent updates may allow stale data to dominate. 

A promising future direction in distributed CL with one replay buffer is the development of an adaptive buffer-sharing mechanism, where \textit{the allocation of buffer space is dynamically tailored to the temporal variability of each client’s data distribution}. This approach could explore whether clients whose data distribution exhibits more pronounced shifts over time would merit a larger share of the centralized replay buffer to capture emerging patterns and mitigate forgetting. This problem would require a method to rigorously quantify in real time the statistical variation of data over time, in order to detect distribution changes, compare their significance, and allocate buffer space accordingly.

\par Additionally, adapting $\mathrm{RS}$-based strategies in a distributed CL setting with one replay buffer per client is not trivial. In this setting, each client independently applies ($\mathrm{RS}$) to curate its buffer contents to the local data stream. A key question is whether the collective outcome of these decentralized $\mathrm{RS}$ processes can be statistically equivalent to performing $\mathrm{RS}$ on a centralized dataset. The challenge would be to find a global $\mathrm{RS}$ policy so that the union of clients' local buffers approximates the unbiased representative sample of the global distribution that one centralized buffer with $\mathrm{RS}$ would yield. 

\subsubsection{Federated Continual Learning}

An interesting future direction in FCL is the design of buffer management signaling strategies that enable clients to make informed decisions about which data samples to retain or evict. Each client could independently optimize its buffer to preserve a diverse and balanced subset of historical data and to mitigate the effects of catastrophic forgetting on its own local data distribution. However, buffer updates across clients would benefit from using a global signal from the server acting as a central coordinator with the goal of achieving model convergence. Such a signal could guide clients in aligning their buffer contents to improve the global model’s generalization performance. 

\section{Conclusion}

We presented a modeling framework for integrating Federated Learning with XAI and Continual Learning at the network edge. Our approach supports distributed training of complex black-box ML models alongside interpretable surrogate ones via a MOO formulation in a way that balances accuracy and explainability. We started from the fundamental model of centralized training of inherently interpretable decision trees and further discussed how such models can be trained in the challenging setting of FL. Moreover, by examining centralized (shared buffer) and decentralized (local buffers) architectures, we provided some insights into addressing CL at the network edge in the presence of non-stationary data distributions.

We believe that the goal of having interpretable and continually updated ML models in distributed settings at the network edge is the cornerstone to having trustworthy and reliable AI-based services at the network edge. These services would be realized by training explainable ML models from continuous streams of data at wirelessly connected and resource-limited edge devices. In this context, we expect that this work will serve as a stepping stone for further developments in this field.

\section*{Acknowledgment}
This work was conducted in the context of the Horizon Europe project PRE-ACT (Prediction of Radiotherapy side effects using explainable AI for patient communication and treatment modification). It was supported by the European Commission through the Horizon Europe Program (Grant Agreement number 101057746), by the Swiss State Secretariat for Education, Research and Innovation (SERI) under contract number 22 00058, and by the UK government (Innovate UK application number 10061955).



\begin{thebibliography}{00}
\bibitem{Xia} Q. Xia, W. Ye, Z. Tao, J. Wu, and Q. Li, ``A survey of federated learning for edge computing: Research problems and solutions,'' \textit{High-Confidence Computing}, vol. 1, no. 1, p. 100008, 2021.
\bibitem{tsoup} T. Tsouparopoulos and I. Koutsopoulos, ``Implementation of Federated Learning on Resource-constrained devices: Lessons learned," In Network Intelligence Workshop of IFIP Networking Conference, Italy, 2022
\bibitem{McMahan} B. McMahan, E. Moore, D. Ramage, S. Hampson, and B. A. y Arcas, ``Communication-efficient learning of deep networks from decentralized data,'' in \textit{Proc. Int. Conf. AISTATS}, USA, 2017, pp. 1273--1282.
\bibitem{Charal} F. Charalampakos and I. Koutsopoulos, ``Exploring multi-task learning for explainability,'' in \textit{Proc. 24th Eur. Conf. Artif. Intell. (ECAI)}, Cham, Switzerland, Sep. 2023, pp. 349--365.

\bibitem{Ribeiro}
M. T. Ribeiro, S. Singh, and C. Guestrin, ``‘Why should I trust you?’: Explaining the predictions of any classifier,'' in \textit{Proc. 22nd ACM SIGKDD Int. Conf. KDD}, San Francisco, USA, 2016, pp. 1135--1144.
\bibitem{Plumb}
G. Plumb, M. Al-Shedivat, Á. A. Cabrera, A. Perer, E. Xing, and A. Talwalkar, ``Regularizing black-box models for improved interpretability,'' in \textit{Proc. 33rd Conf. NeurIPS}, Canada, 2020, pp. 10526--10536.

\bibitem{Fei} Z. Fei, B. Li, and Y. Zhang, ``A survey of multi-objective optimization in wireless sensor networks,'' \textit{arXiv preprint arXiv:1609.04069}, 2016.

\bibitem{Désidéri} J.-A. Désidéri, ``Multiple-gradient descent algorithm for multi-objective optimization,'' \textit{J. Optim. Theory Appl.}, pp. 491--509, 2012.

\bibitem{Sener} O. Sener and V. Koltun, ``Multi-task learning as multi-objective optimization,'' in \textit{Proc. 32nd Conf. NeurIPS}, Canada, 2018, pp. 527--538.
\bibitem{Hu} Y. Hu, X. Yuan, and Y. Huang, ``Federated multi-objective learning,'' in \textit{Proc. 36th AAAI Conf. }, Virtual, 2022, pp. 7845--7853.
\bibitem{Yang} H. Yang, Z. Liu, J. Liu, C. Dong, and M. Momma, ``Federated multi-objective learning,'' in \textit{Proc. 36th Conf. NeurIPS}, USA, 2024.

\bibitem{QLi} Q. Li, Z. Wen, and B. He, ``Federated learning systems: Vision, hype and reality for data privacy and protection,'' \textit{arXiv preprint arXiv:1907.09693}, 2019.
\bibitem{Ludwig} H. Ludwig et al., ``IBM federated learning: An enterprise framework white paper $v0.1$,'' \textit{arXiv preprint arXiv:2007.10987}, 2020. [Online]. 

\bibitem{Patton} Q. Li, Z. Wen, and B. He, ``Practical federated gradient boosting decision trees,” in Proc. AAAI Conf. Artif. Intell., vol. 34, no. 04, 2020.

\bibitem{Argente-Garrido} A. Argente-Garrido, C. Zuheros, M. Luzón, and F. Herrera, ``An interpretable client decision tree aggregation process for federated learning,'' \textit{arXiv preprint arXiv:2404.02510}, 2024.

\bibitem{Aljundi} R. Aljundi, K. Kelchtermans, and T. Tuytelaars, ``Task-free continual learning," in Proc. IEEE/CVF Conf. Comput. Vis. Pattern Recognit., 2019, pp. 11254-11263.
\bibitem{Slin} S. Lin, L. Yang, D. Fan, and J. Zhang, ``Beyond Not-Forgetting: Continual Learning with Backward Knowledge Transfer," in Advances in Neural Information Processing Systems 35 (NeurIPS 2022), 2022.
\bibitem{Yoon20} J. Yoon, S. Kim, E. Yang, and S. J. Hwang, ``Scalable and order-robust continual learning with additive parameter decomposition," in Eighth International Conference on Learning Representations (ICLR), 2020.
\bibitem{Yoon} J. Yoon, E. Yang, J. Lee, and S. J. Hwang, ``Lifelong learning with dynamically expandable networks," arXiv preprint arXiv:1708.01547, 2017.
\bibitem{Vitter} J. S. Vitter, ``Random sampling with a reservoir,'' \textit{ACM Trans. Math. Softw.}, vol. 11, no. 1, pp. 37--57, Mar. 1985.
\bibitem{Chrysakis} A. Chrysakis and M.-F. Moens, ``Online continual learning from imbalanced data,'' in \textit{Proc. ICML}, Austria, 2020, pp. 1952--1961.
\bibitem{Nikol24} S. Nikoloutsopoulos, I. Koutsopoulos, and M. K. Titsias, “Kullback-Leibler Reservoir Sampling for Fairness in Continual Learning,” in Proc. IEEE ICMLCN, 2024, pp. 460–466.


\bibitem{Lopez-Paz} D. Lopez-Paz and M. Ranzato, ``Gradient episodic memory for continual learning,'' in \textit{Proc. 31st NeurIPS}, USA, 2017, pp. 6470--6479.

\bibitem{Charal25}  F. Charalampakos, T. Tsouparopoulos and I. Koutsopoulos, “Joint Explainability-Performance Optimization with Surrogate Models for AI-Driven Edge Services,” to appear, IEEE Int. Conf. on Machine Learning for Commun. and Networking (ICMLCN) 2025.
\bibitem{Rokach} O. Z. Maimon and L. Rokach, ``Data Mining With Decision Trees: Theory and Applications, 2nd ed. Singapore: World Scientific, 2014.
\bibitem{Yoon21} J. Yoon, W. Jeong, G. Lee, E. Yang, and S. J. Hwang, “Federated continual learning with weighted inter-client transfer,” in Proc. Int. Conf. Mach. Learn., Jul. 2021, pp. 12073–12086.



\end{thebibliography}
\end{document}